\documentclass[letterpaper, 10 pt, conference]{ieeeconf}  

\IEEEoverridecommandlockouts                              

\usepackage{color}
\usepackage{graphicx}
\usepackage{amsmath}
\usepackage{diagbox}
\usepackage{booktabs}
\usepackage{subfigure}
\usepackage{cite}

\title{\LARGE \bf
Traffic Agent Trajectory Prediction Using Social Convolution and Attention Mechanism
}

\author{Tao Yang, Zhixiong Nan*, He Zhang, Shitao Chen and Nanning Zheng
\thanks{*Corresponding author: Zhixiong Nan{ \tt\small nzx2018@xjtu.edu.cn}}
\thanks{The authors are with the Institute of Artificial Intelligence and Robotics, Xi'an Jiaotong University, Xi’an, China}
}

\begin{document}

\maketitle
\thispagestyle{empty}
\pagestyle{empty}

\begin{abstract}
The trajectory prediction is significant for the decision-making of autonomous driving vehicles. In this paper, we propose a model to predict the trajectories of target agents around an autonomous vehicle. The main idea of our method is considering the history trajectories of the target agent and the influence of surrounding agents on the target agent. To this end, we encode the target agent history trajectories as an attention mask and construct a social map to encode the interactive relationship between the target agent and its surrounding agents. Given a trajectory sequence, the LSTM networks are firstly utilized to extract the features for all agents, based on which the attention mask and social map are formed. Then, the attention mask and social map are fused to get the fusion feature map, which is processed by the social convolution to obtain a fusion feature representation. Finally, this fusion feature is taken as the input of a variable-length LSTM to predict the trajectory of the target agent. We note that the variable-length LSTM enables our model to handle the case that the number of agents in the sensing scope is highly dynamic in traffic scenes. To verify the effectiveness of our method, we widely compare with several methods on a public dataset, achieving a 20\% error decrease. In addition, the model satisfies the real-time requirement with the 32 fps.


\end{abstract}

\section{INTRODUCTION}
Benefitting from the advanced sensors such as the laser radar, camera, millimeter-wave radar and equipments as well as the complex processing algorithms, the autonomous vehicles can accurately perceive the surrounding environment\cite{chen2019autonomous}. Based on perception results, the planning and control algorithm is able to control the autonomous vehicles to follow the specified route and avoid collisions. However, in some complex driving scenes, this mode might lead to serious consequences such as the traffic accidents of Tesla and Uber in 2018. These phenomena result from the lack of predictability of the future trajectory of agents in the planning algorithm. Experts believe that autonomous vehicles with the predictability of the future trajectory of agents can avoid similar accidents\cite{zhan2018probabilistic}.


\begin{figure}[!htbp]
\centering
\subfigure[Trajectory prediction results under egocentric vision]{
\begin{minipage}[t]{1.0\linewidth}
\centering
\includegraphics[width=3in]{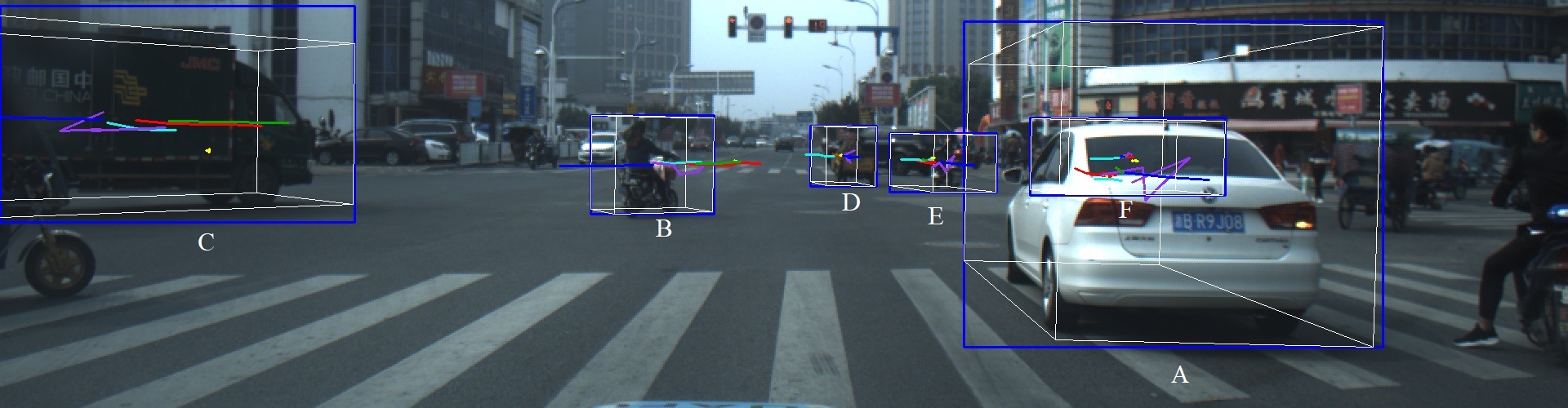}
\label{fig:1}
\end{minipage}%
}%

\subfigure[Trajectory prediction results under radar map]{
\begin{minipage}[t]{1.0\linewidth}
\centering
\includegraphics[width=3in]{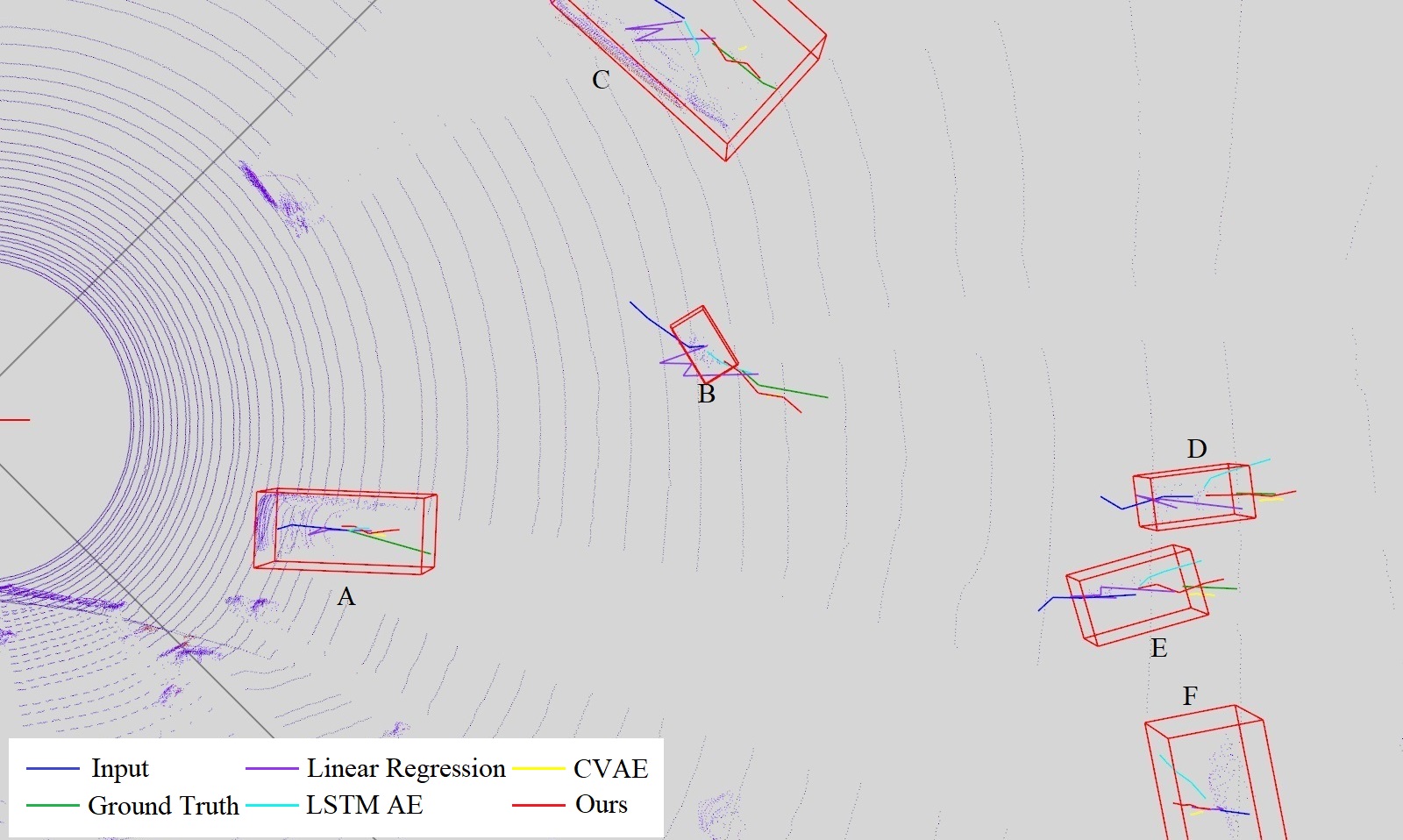}
\label{fig:2}
\end{minipage}%
}%
\centering

\caption{Trajectory Prediction in High-Density Traffic Scene. The red cuboids in (b) from big to small are vehicles, riders, and pedestrians, respectively. The circles on the left represent autonomous vehicles. In (a) and (b), the blue lines are inputs, the green lines are Ground Truth,  The blue lines are inputs, the green lines are ground truth, the purple, blue, yellow and red lines are the prediction results of Linear Regression, LSTM AE, CVAE and our model respectively}
\label{figure:000}
\vspace{-0.2in}
\end{figure}

The factors that affect agent trajectories in traffic scenarios are particularly numerous and complex, therefore, trajectory prediction in the autonomous driving scenario is an extremely challenging task. These factors include the type of the agent (pedestrian, rider, vehicle) \cite{ma2019trafficpredict}, the traffic rules\cite{lee2017desire}, interaction between the different type of agents\cite{li2019grip}, the drivers’ subjective decision\cite{deo2018multi} etc. Some early works that focus on different aspects of these factors have been proposed. However, these works didn't model the regression problem from the perspective of an agent's decision making, which allows us consider these factors in a more intuitive way and decrease the computation consumption. Some other works \cite{li2019grip, alahi2016social, yao2019egocentric} regard the regression problem as a regression of the traffic scene, which makes it difficult to deal with the agent entering and leaving of the scope of the sensing scope. 
One recently-proposed work \cite{yao2019egocentric} uses the egocentric visual cue and the optical flow to predict the future bounding box of the agent, but the vision-based trajectory prediction is not accurate. The work in \cite{luo2018fast} uses 3D point cloud computing for simultaneous detection, tracking, and trajectory prediction, but it requires a large amount of calculation.

In this paper, inspired by the above-mentioned methods, we propose a trajectory prediction model that involves an attention mechanism and a social map, targeting to consider the procedure of the decision making and decrease the computation consumption. When making a decision, the ego-vehicle usually pays more attention to surrounding agents and less attention to distant agents. Motivated by this observation, we encode the history trajectories of the target agent as an attention mask and the positions of surrounding agents as a social map. The attention mask is actually a probability map with high probabilities at the regions around the target agent and low probabilities at the regions far from the target agent. This attention mask is fused with the social map to encode the importance of surrounding agents. In addition, this mechanism allows decreasing the computation consumption since the trajectory prediction is conducted only utilizing surrounding agents instead of all agents in the traffic scene. Given a trajectory sequence as input, the attention mask and social map are firstly formed based on the original LSTM features of agents. Then, the attention mask and social map are fused and processed a social convolution to output a fusion feature. Finally, the fusion feature, together with the original LSTM feature, are concatenated to serve as the input of a variable-length LSTM to predict the trajectory of the target agent. We note that the variable-length LSTM enables our model to handle the case that the number of agents in the sensing scope is highly dynamic in traffic scenes. In the experiments, our method is compared with several methods, achieving the best performance on three different metrics. In addition, the ablation study experiments are conducted to verify the effectiveness of our attention mechanism and social convolution.



The contributions of this paper are as follows: 
\begin{itemize}
  \item An efficient and accurate trajectory prediction framework is proposed to improve the trajectory prediction accuracy of traffic agents around autonomous vehicles.
  \item The attention mechanism and social map are proposed to consider the procedure of decision making and decrease the computation consumption.
\end{itemize}

The rest of this paper is organized in the following order. We discuss related works in Section II, followed by the problem formulation in Section III. In Section IV, we detail our method. We present the implementation details and report our experimental results in Section V. Finally, we conclude this paper in Section VI.

\section{RELATED WORK}

Trajectory prediction has been researched extensively. Traditional methods include the the Bayesian formulation \cite{lefevre2011exploiting}, Hidden Markov Models (HMMs) \cite{firl2012predictive}, Kalman Filters \cite{kalman1960new}, the Monte Carlo simulation\cite{danielsson2007monte}, the Gaussian processes\cite{laugier2011probabilistic} and LSTM autoencoders \cite{park2018sequence}. These traditional methods did not take the complex interactions between agents into consideration. Thus, here we merely summarize the more recent works that take the complex interactions between agents into account.

Alahi et al. proposed a social pooling layer in LSTM to extract surrounding (local) agents' information to help the trajectory prediction and called Social LSTM\cite{alahi2016social}, it is the first one proposed to use scene information to assist trajectory prediction, and it can make the precise prediction. However, since it predicts the future trajectory of the whole traffic scene, which makes it difficult to handle the frequent entry and exit of agents in the scene. Besides, in every time step, it needs to compute the social pooling of the agents, therefore, the amount of calculation is large.

Deo et al. proposed a convolution social pooling layers to fuse the surrounding agents (global) information \cite{deo2018convolutional}. It divides one lane into cells, different lanes form a grid. It uses the convolutional neural network to fuse the information of the agents located in different cells, which can solve the problem of frequent entry and exit of agents in the scene. And social convolution fully utilized the location information of the agents. However, the prediction is made according to the maneuver classification, so that the maneuver prediction error has a great impact on the trajectory prediction, and the type of agents is not taken into account. 

Li et al. proposed a graph convolutional model to model interactions between the agents which is called GRIP \cite{li2019grip}. They regard the agents as nodes and regard the interactive event as the edges, then use graph convolution to extract the effect of the interactive events on the trajectory prediction, it achieves a precise prediction. Since it did not directly utilize the position information of the agents, but take convolution on the original tensor, it takes more effort to learn the interactive events. In addition, it needs to compute the inverse of the matrix, it consumes a large amount of calculation.

\section{PROBLEM FORMULATION}
For the convenience of describing the method, we would like to formulate the trajectory prediction problem before presenting our model. The observable scene of the autonomous vehicle at time $t$ is denoted as $s^{(t)}$, thus the input of our model $X$ are the historical scenes over $t_h$ time steps:
\begin{equation}
X=\left[s^{(1)}, s^{(2)} \cdots, s^{\left(t_{h}\right)}\right]
\end{equation}
when the autonomous vehicle moves, there are agents entering and leaving the observable region of it. The number of agents at different times might be different, suppose that there are $n_t$ agents in the observable region at time $t$. Thus, the observable scene at time $t$ is:
\begin{equation}
s^{(t)}=\left[p_{1}^{(t)}, p_{2}^{(t)},\cdots, p_{n_t}^{(t)}\right]
\end{equation}
where $p_n^{(t)}, n = 1,2,\cdots, n_t$ is the position of agent $n$ at time $t$, considering that there are slopes during driving and different lanes may have different heights sometimes. We believe that it is necessary that our coordinate includes z-axis, the coordinate in this condition is:
\begin{equation}
p_{n}^{(t)}=\left[x_{n}^{(t)}, y_{n}^{(t)}, z_{n}^{(t)}\right]
\end{equation}
the coordinate used here is the ego-vehicle-based coordinate system with relative measurement, in the above context, assume that the model needs to predict the trajectory of the agents from time step $t_h + 1$ to $t_h + t_f$, the output of the model of time $t_h$ is as follows:
\begin{equation}
Y=\left[s^{\left(t_{h}+1\right)}, s^{\left(t_{h}+2\right)}, \cdots, s^{\left(t_{h}+t_{f}\right)}\right]
\end{equation}
where the definition of $s^{(t)}$ is the same as the input, however, the number of the agents is as same as the time step $t_h$. 
\section{APPROACH}
\begin{figure*}[htbp!]
\centering
\vspace{0.06in}
\includegraphics[width=13.5cm]{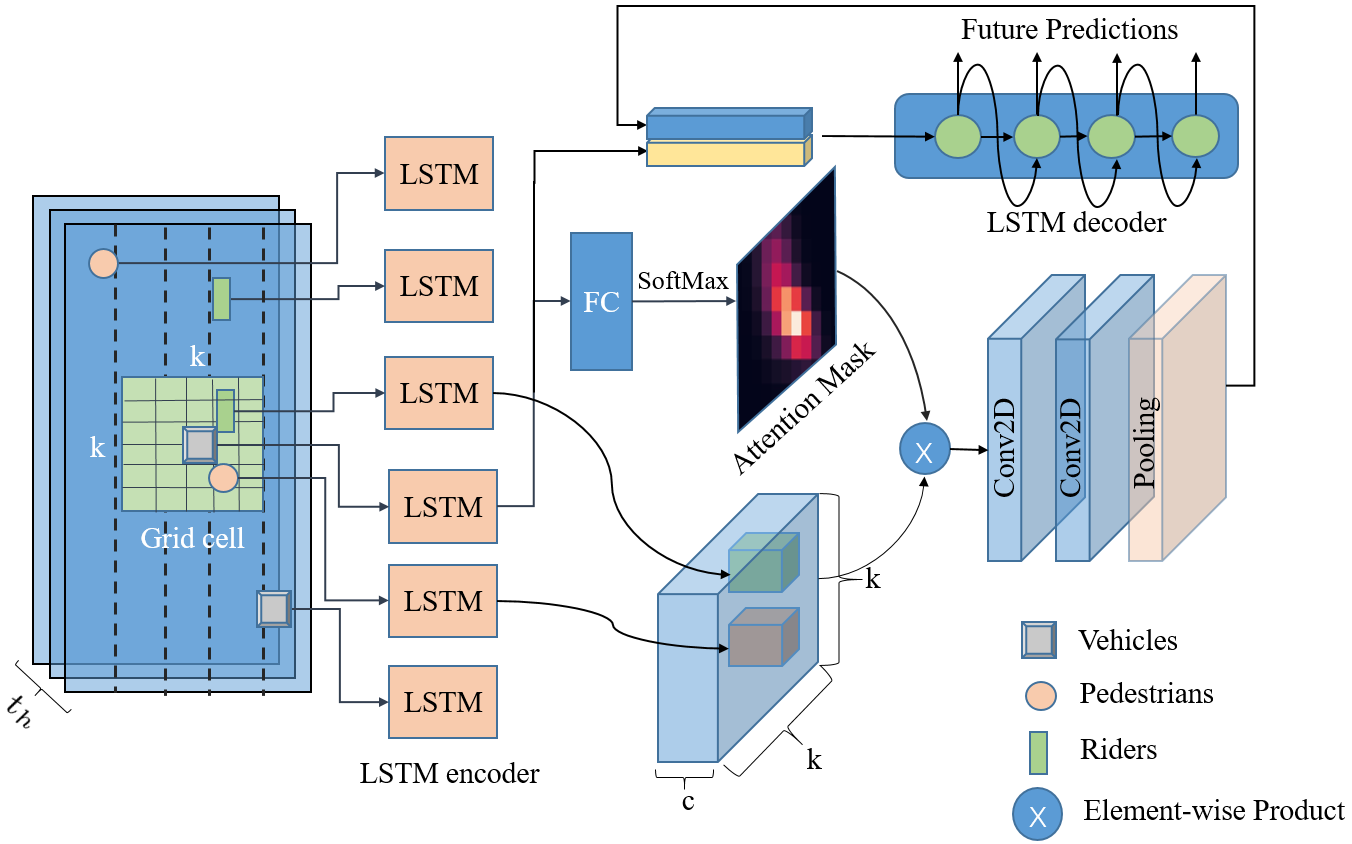}
\caption{The architecture of the proposed approach: The target agent is marked by the grey square. The blue grid region around it is its grid cell. We generate input representation for all agents based on trajectory information. These representation are passed through LSTMs and eventually used to construct the social map, the target agent's representation is encoded as the attention mask. The production of attention mask and social map is passed through ConvNets and then concatenated together with the target agent tensor to produce latent representation. Finally, this latent representation are passed through an LSTM to generate a trajectory prediction for the target agent.}
\label{fig:picture001}
\vspace{-0.06in}
\end{figure*}
We divide the proposed deep neural network model into four parts: input representation module, LSTM encoder module, attention mask, and social map fusion and LSTM decoder module. The overall architecture is shown in Fig. \ref{fig:picture001}. The raw data is processed and converted to structured data by using the part 1, and then we extract the historical trajectory representation of all the agents by using the part 2, we fuse the surrounding agents' representation to obtain a social representation by applying the part 3, this feature representation is concatenated with history trajectory feature representation to serve as the input of part 4 to finally obtain the future prediction.

\subsection{Input Representation Module}
The complex traffic scene leads to the unstructured raw data, before feed it to our model, we need to transform the raw data into structured data.

We consider a scenario with time $ t $, we need to predict $ n $ agents. When predicting the trajectory of the $i$-th agent by modeling its planning, assuming there are $n_i$ agents in the $m \times m$ size observable region, we divide the $m \times m$ observable region into $k \times k$ grid. Then label the agent in the corresponding grid cell, represent the historical trajectory of the predicted agent as a tensor with a size of $(b,t_h,h)$, where $b$ denote the batch size. We set $h=3$ to indicate $x,y$ and $z$ coordinates of the agent. 

As shown in Fig. \ref{fig:picture001}, we represent the grid as a tensor with a size of $(k,k,1)$, set the cell with the agent in it as its label, set the value of other cells $0$. Because there are different number of agents in the observable region, we represent them by a dictionary: \{Agent\_label: $(t_1,c_1),\dots,(t_{n_i},c_{n_i})$\}. However, this representation can not feed into our model directly, we need to pad it into a tensor with size of $(b,n_b,t_b,c)$, where $n_b$ represents the maximum number of agents in the grid of the batch, and $t_b$ represents the maximum time length of the historical trajectory of the agents in the batch. It is worth noting that the tensor with a size of $(b,n_b,t_b,c)$ is used as the input of the LSTA Encoder Module and the tensor with a size of $(k,k,1)$ is used to form the social map.

\subsection{LSTM Encoder Module} To encode the historical trajectory information, we feed the trajectories of the different agents with the different time lengths into a dynamic LSTM, taking the hidden state of the corresponding time length as the representation of the trajectory. The representation is concatenated with the class label's one-hot representation to be the representation $z_i$.

\subsection{Attention Mask and Social Map Fusion}
\subsubsection{Social Map} We take out the historical trajectory representation $z_i,i=1,2\dots,n_j$, which is in the grid of the agent $j$. Then we plug it into the corresponding position of the grid cell by using the label we defined in the input representation module, forming a tensor of size $(k,k,c)$ as Fig. \ref{fig:picture001} shows. In this case, only those grids containing agents contain the representation of their historical trajectories, while all the other positions are 0. This tensor contains the location information and historical trajectory information, we represent the tensor as $G_i$.

\subsubsection{Trajectory-based Attention Mask} In driving scenario, the agent at the time of decision making, will pay attention to those agents that enter their observable zone and could be dangerous, thus when using the surrounding information to do auxiliary trajectory prediction, not every agent is equally important, the importance is related to the centric agent's historical trajectory, and also other agents' positions. Therefore, we model this process \cite{bahdanau2014neural} by feeding the trajectory representation of the centric agent into a fully-connected network to predict a grid mask $M$:
\begin{equation}
M = softmax(FC(z_i))
\end{equation}
where $z_i$ is the trajectory representation of the centric agent derived in the  LSTM encode module, the value of $M$ is in $[0,1]$ indicating how important the agent is for predicting the trajectory of the centric agent. 
\subsubsection{Social Convolution Fusion} We let the derived social map $G_i$ and the attention mask $M$ do element-wise product, then feed it into a convolutional layer with ReLU as the activation to compute convolutional feature maps, which is further processed by a convolutional layer to fuse the information of different positions. Finally it is fed into a max-pooling layer to further extract the surrounding information as Fig. \ref{fig:picture001} shows. Then a fully-connected network  is used to embed the social map into the same representation space with $z_i$, and they are concatenated to be a trajectory representation with agents' interactions.

\subsection{LSTM Decoder Module} This module predicts the centric agent trajectory by taking the trajectory representation with agents' interactions as input. As Fig. \ref{fig:picture001} shows, we take it as the initial hidden state of the LSTM decoder, predicting trajectory by outputting a vector $(b,t_h, c)$, where $c$ varies according to the type of Loss function we use. In details, if we use L2 Loss as the loss function, $c = 3$, which represents x, y, z. If the Loss function GMM Loss, $c = 7$, which represents $x_m, y_m, x_\sigma, y_\sigma, \rho, z$.

\subsection{Loss Function}
We use L2 Loss to regress the prediction coordinates, then the overall loss can be computed as:
\begin{equation}
\begin{aligned}L o s s &=\frac{1}{t_{f}} \sum_{t=1}^{t_{f}}\left\|Y_{p}^{t}-Y_{GT}^{t}\right\|^{2} \\&=\frac{1}{t_{f}} \sum_{t=1}^{t_{f}}\frac{1}{n_t} \sum_{i=1}^{n_t}\left\|p_{i_p}^{t}-p_{i_{GT}}^{t}\right\|^{2}\end{aligned}
\end{equation}
where $t_f$ is the prediction time length, $Y_{p}$ denote the prediction coordinate of the agents, $Y_{GT}$ is the prediction ground truth.
We regard the model as a probability density estimation model, and use the gaussian mixture model to model the prediction trajectory probability. In this case, the objective function is:
\begin{equation}
\theta = \mathop{\arg\max}_{\theta} \log P(Y_p|X,\theta)
\end{equation}
where $\theta$ denotes the model parameters, $X$ represents the input historical scenes, and $Y_p$ denotes the predicted scenes. In this case, the loss function is GMM Loss.

\begin{table*}
\caption{The results for trajectory prediction on BLVD dataset.}
\begin{tabular*}{\hsize}{@{}@{\extracolsep{\fill}}ccccccccccccc@{}}
\toprule
Dataset                     & \multicolumn{3}{c}{all}             & \multicolumn{3}{c}{pedestrian}      & \multicolumn{3}{c}{vehicle}         & \multicolumn{3}{c}{rider}           \\
\cmidrule(r){1-1} \cmidrule(r){2-4} \cmidrule(r){5-7} \cmidrule(r){8-10}\cmidrule(r){11-13}
\diagbox{Model}{Metric} & ADE & MDE & FDE & ADE & MDE & FDE & ADE & MDE & FDE & ADE & MDE & FDE \\
\midrule
Linear Regression           & 3.31          & 3.77          & 3.35          & 2.68          & 3.03          & 2.69          & 3.31          & 3.77          & 3.34          & 3.04          & 3.53          & 3.09          \\
LSTM\_AE\cite{graves2013generating}                    & 1.11          & 1.69          & 1.60          & 0.94          & 1.43          & 1.35          & 1.12          & 1.71          & 1.63          & 1.12          & 1.75          & 1.64          \\
CVAE\cite{sohn2015learning}                        & 0.81          & 1.28          & 1.05          & 0.70          & 1.17          & 1.01          & 1.23          & 1.27          & 1.03          & 0.82          & 1.32          & 1.11          \\
Ours(L2 Loss)                    & 0.71          & 1.15          & 1.04          & 0.69          & 1.12          & 1.02          & 0.71          & 1.14          & 1.03          & 0.80          & 1.29          & 1.19          \\
Ours(GMM Loss)               & \textbf{0.65} & \textbf{1.04} & \textbf{0.93} & \textbf{0.64} & \textbf{1.01} & \textbf{0.91} & \textbf{0.65} & \textbf{1.04} & \textbf{0.93} & \textbf{0.72} & \textbf{1.15} & \textbf{1.02}\\
\bottomrule
\label{tab1}
\end{tabular*}
\vspace{-0.15in}
\end{table*}

\section{EXPERIMENTS}
\subsection{Settings}
\textbf{Dataset}
The BLVD dataset\cite{xue2019blvd} consists of 654 high-resolution video clips with a total of 120k frames, was extracted from Changshu city, Jiangsu province. 
This dataset includes 6,004 valid event fragments of surrounding participants. In each frame, the ID, 3D coordinates, direction information and the interaction behavior of all objects are recorded. We follow Xue et al.\cite{xue2019blvd} to divide the dataset into the training set and the test set. From four datasets (day high density , day low density, night high density, night low density), here different lighting conditions during the day and night will affect the detection of the agents. 

\textbf{Implementation Details}
We run our model on a desktop running Ubuntu 16.04 with 4.0GHz Intel Core i7 CPU, 32GB Memory, and an NVIDIA Tesla V100 Graphics Card. Our model is implemented by using Python and PyTorch. 

\textbf{hyper-parameters setting} 
We set m to be 30m and grid size k to be 11. The dimension of the output representation is 20, included 17-dimensional historical trajectory information representation and 3-dimensional representation of the agent category, there are three types of agents: vehicles, pedestrians and riders. Hyperparameters of the two convolutional layers are: kernel size: 3, 5, the stride: 2, 2, output channels: 64, 16. The kernel size of the pooling layer is 2. We use Adam optimizer to train the model, set the learning rate to 0.001, take every 10 epochs and multiply the learning rate by 0.1 to decrease until convergence, and set the batch size to 256.

\subsection{Metrics}

Following the metrics used in\cite{lee2017desire} and \cite{yao2019egocentric}. In this paper, we use the following three evaluation metrics to comprehensively measure the performance of the model: Average Displacement Error (ADE), the average displacement error reflects the average level of the prediction error, which can be calculated by the following formula:
\begin{equation}
ADE=\frac{1}{n} \sum_{i=1}^{n}\frac{1}{t_{f}} \sum_{t=1}^{t_{f}}\left\|p_{i_p}^{t}-p_{i_{GT}}^{t}\right\|
\end{equation}
Maximum Displacement Error (MDE): the upper bound of the prediction error, which can be calculated by the following formula:
\begin{equation}
MDE= \frac{1}{n} \sum_{i=1}^{n}\max_{t=1,\dots,t_f}\left\|p_{i_p}^{t}-p_{i_{GT}}^{t}\right\|
\end{equation}
Final Displacement Error (FDE): the displacement error of the predicted trajectory's final point, which can be calculated by the following formula:
\begin{equation}
FDE=\frac{1}{n} \sum_{i=1}^{n}\left\|p_{i_p}^{t_f}-p_{i_{GT}}^{t_f}\right\|
\end{equation}
\subsection{Comparision Results}
In this subsection, to verify the effectiveness of our model, we compare our model with three baseline methods, which are briefly introduced as follows.
\begin{itemize}
  \item \textbf{Linear Regression(LR)} estimates linear parameters by minimizing the least square error. 
  \item \textbf{LSTM autoencoder(LSTM AE)}\cite{graves2013generating} takes the historical trajectory as the input to extract the intention representation and takes it as the hidden state of the LSTM decoder to predict the future trajectory.
  \item \textbf{Conditional VAE(CVAE)}\cite{sohn2015learning} uses variational autoencoder as the model which takes the historical trajectory as the input of encoder, concatenate the one-hot representation to the output of the encoder and takes it as the input of the decoder to get the future prediction. 
\end{itemize}

The comparison results are reported in Tab. \ref{tab1}, from which we observe our model significantly outperforms the baselines on all of the datasets, especially in the vehicle dataset, which has the most samples. We analyze our experimental results from the following two aspects:

\textbf{Baseline} LSTM\_AE performs significantly better than Linear Regression since it can learn the non-linear motions. We observe that CVAE performs better than LSTM\_AE on all of the datasets since CVAE takes advantage of the information on the types of agents. Both of them are not accurate in predicting the trajectory of the vehicle dataset than the other two datasets, because the vehicle dataset has more samples than the other two datasets and the trajectories are more complex and less predictable.
\begin{figure}[t]
\centering
\subfigure[Day high-density]{
\begin{minipage}[t]{0.5\linewidth}
\centering
\includegraphics[width=1.6in]{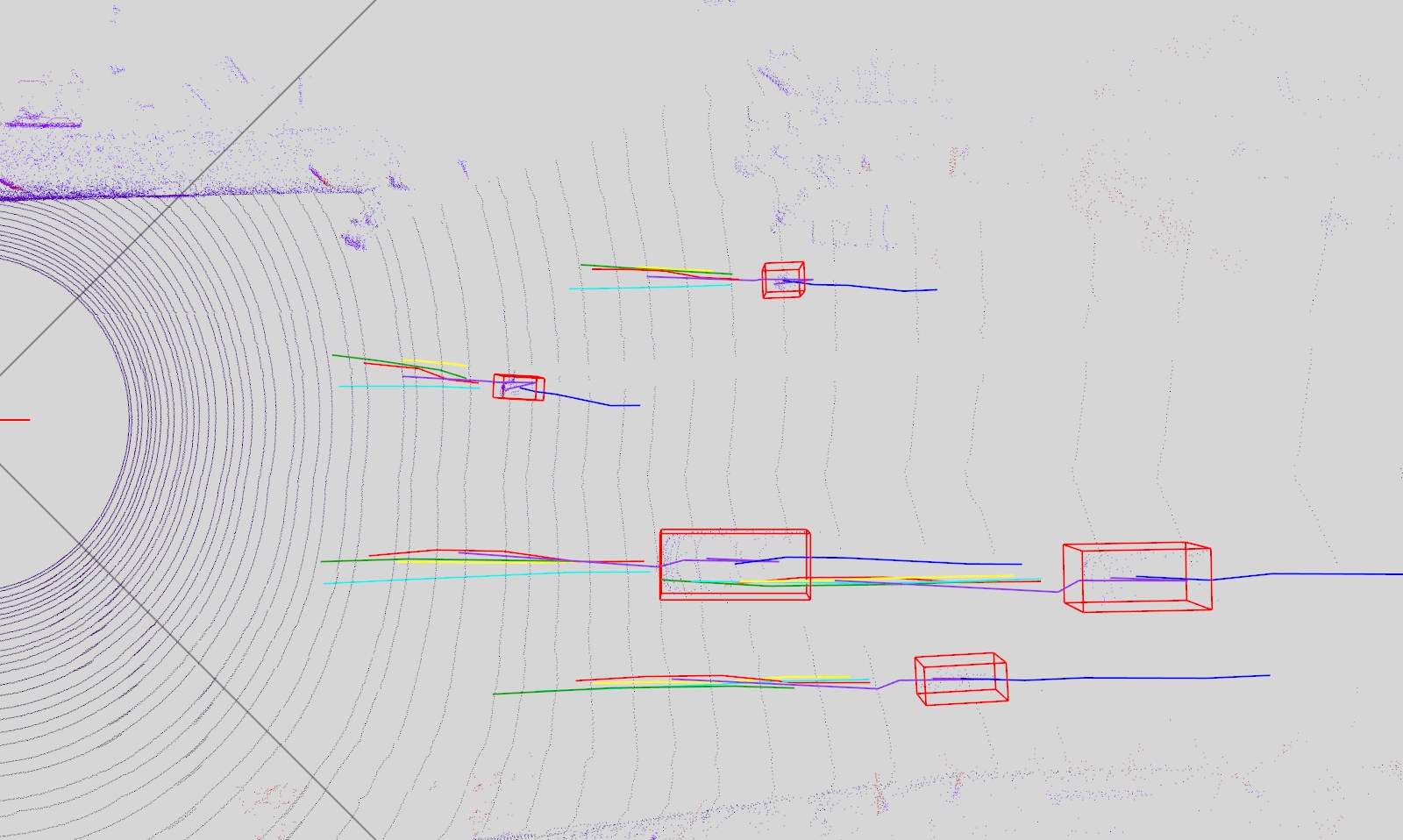}
\label{fig:picture002}
\end{minipage}%
}%
\subfigure[Day low-density]{
\begin{minipage}[t]{0.5\linewidth}
\centering
\includegraphics[width=1.6in]{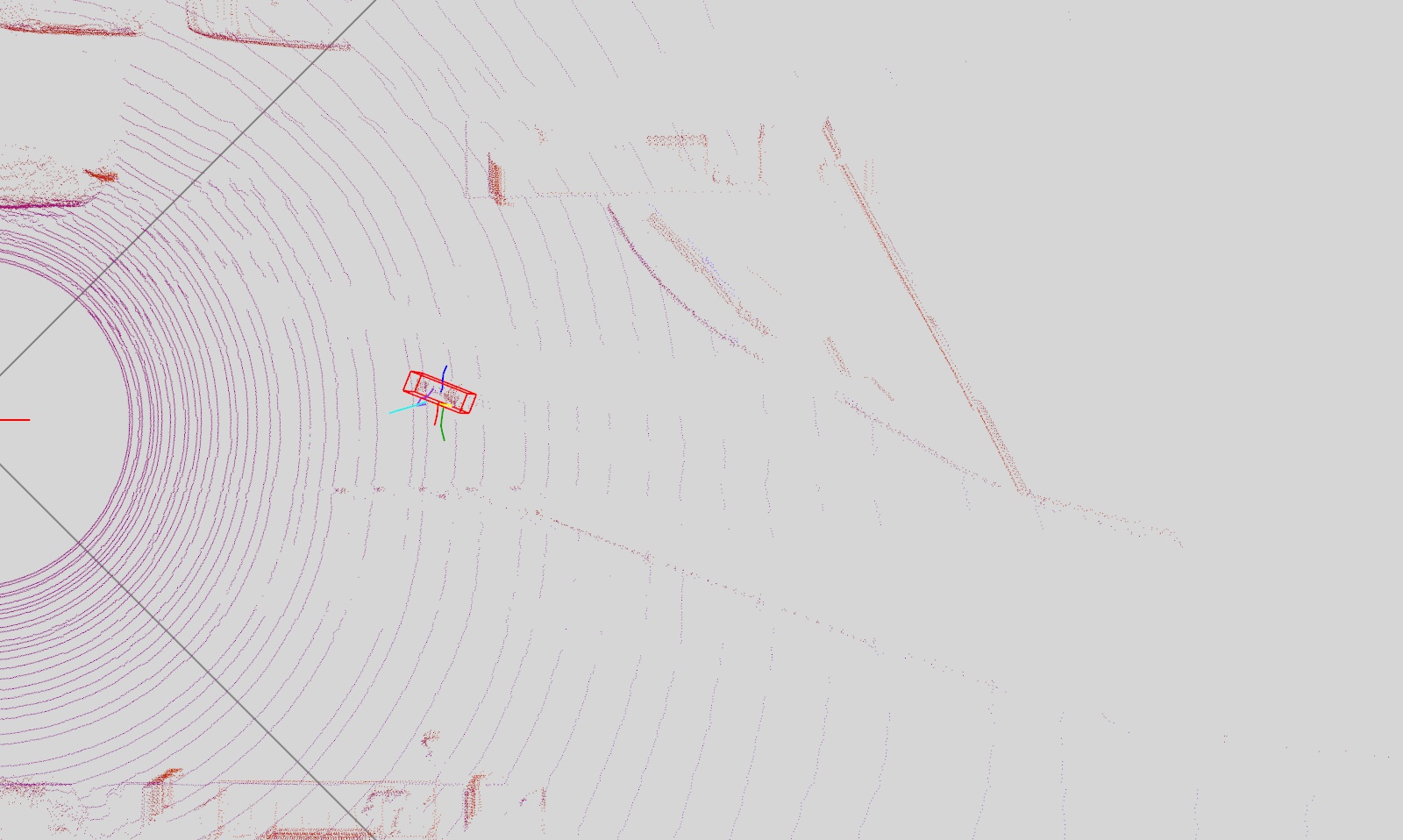}
\label{fig:picture003}
\end{minipage}%
}%
\centering

\centering
\subfigure[Night high-density]{
\begin{minipage}[t]{0.5\linewidth}
\centering
\includegraphics[width=1.6in]{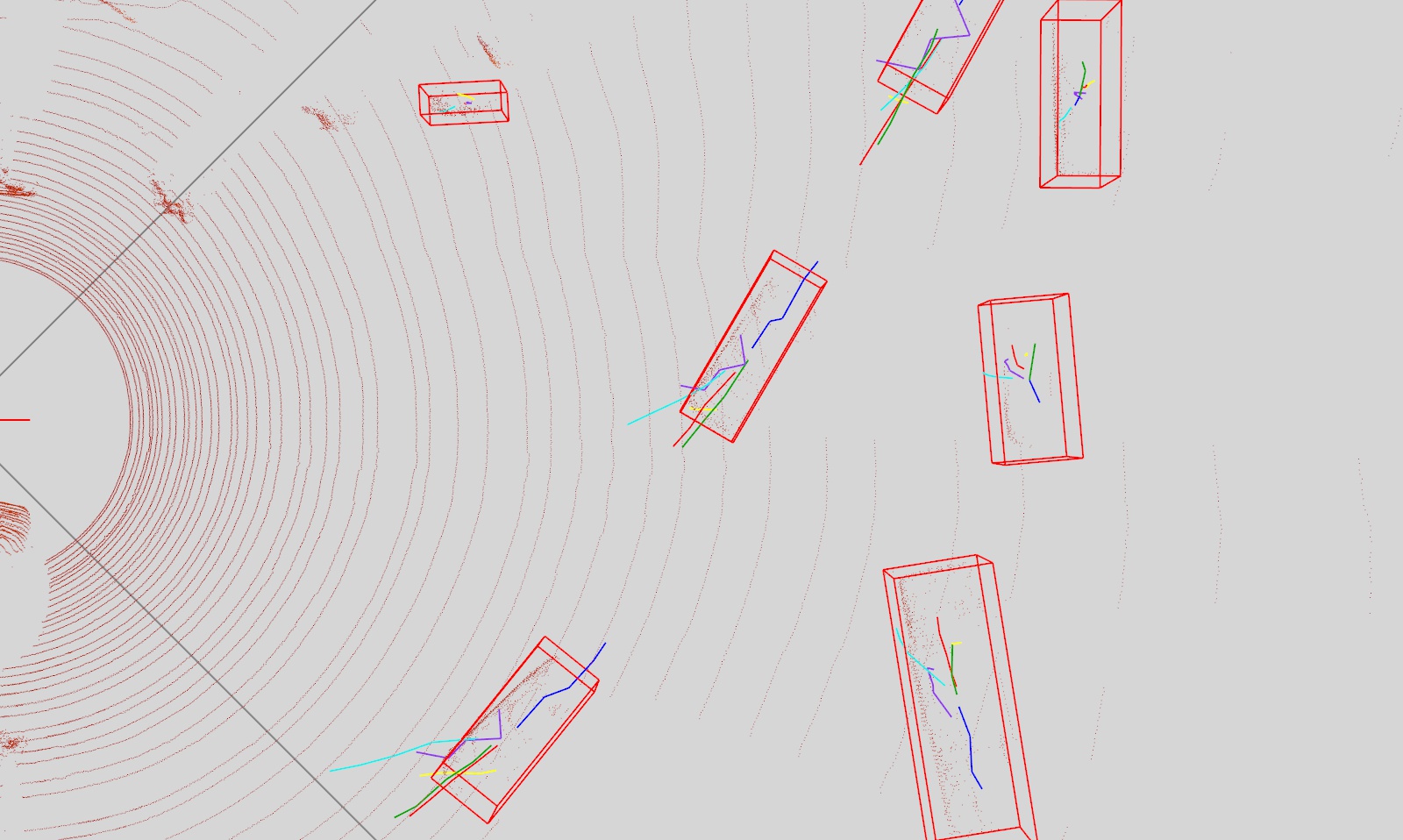}
\end{minipage}%
}%
\subfigure[Night low-density]{
\begin{minipage}[t]{0.5\linewidth}
\centering
\includegraphics[width=1.6in]{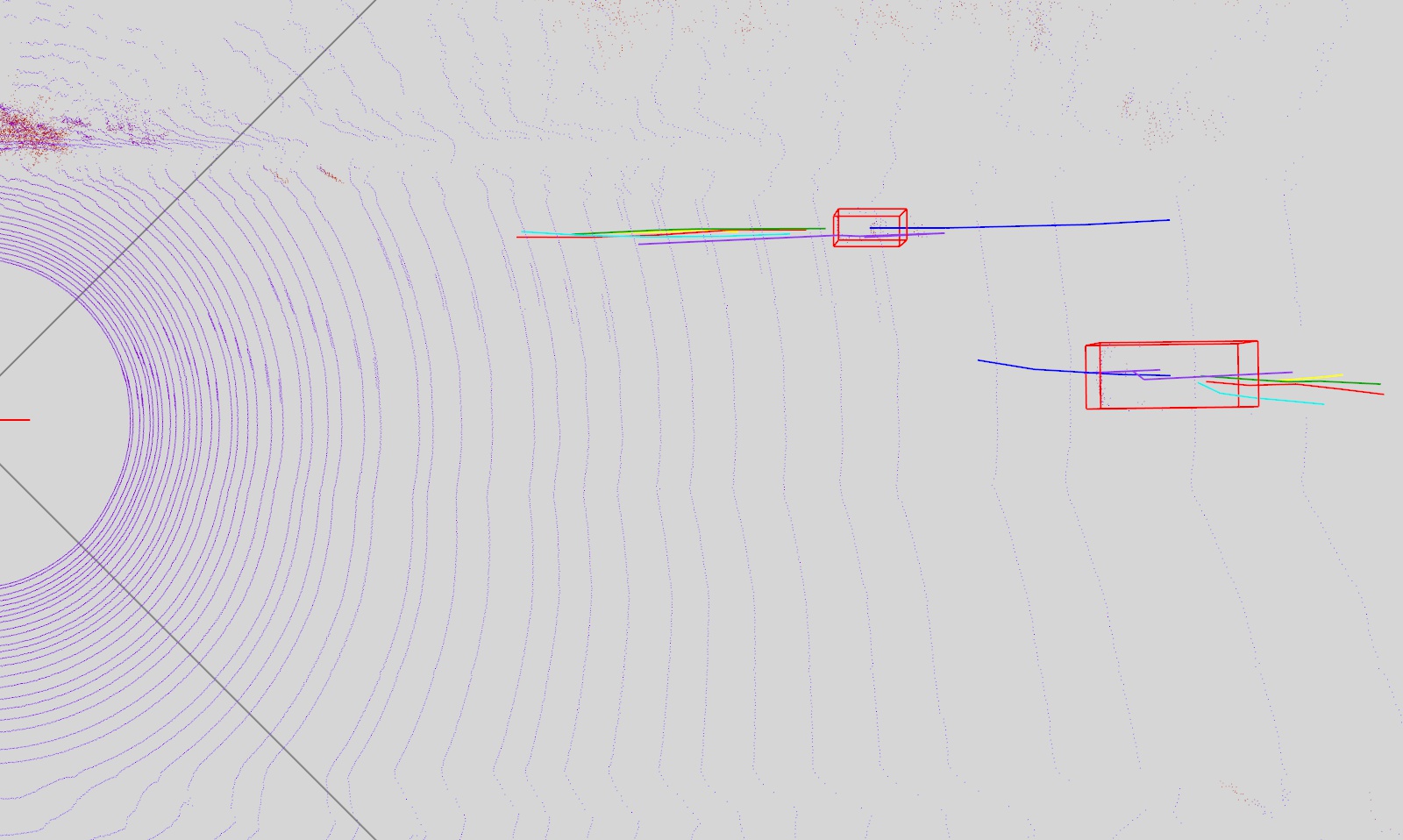}
\end{minipage}%
}%
\centering
\caption{Visualized prediction results. The red cuboids from big to small are vehicles, riders, and pedestrians, respectively. The circles on the left represent autonomous vehicles. The blue lines are inputs, the green lines are ground truth, the purple, blue, yellow and red lines are the prediction results of Linear Regression, LSTM AE, CVAE and our model respectively}
\end{figure}

\textbf{Proposed models} Our model with L2 Loss outperforms the CVAE  by 12\%. Since on the one hand, it utilizes the information of surrounding agents to assist the trajectory prediction, on the other hand, we use the centric agent intention to predict an attention mask to emphasize some of the important information of surrounding agents. Besides, our model with GMM Loss outperforms the CVAE  by 20\%, since GMM loss predicts the output's distribution rather than the trajectory itself, leading to the information extracted from the surroundings is more accurate, which makes our attention mask more accurate. In addition, the poor performance of our model in riders dataset is due to the rider's weak dependence on surrounding information. What's more, we measure computing speed of our proposed model. In the testing process, it runs 32 fps, which outperforms the model without attention that runs 12 fps.

\subsection{Qualitative Results}
We show the visualization results of trajectory prediction in four different scenarios in Fig. \ref{fig:picture002}  and Fig. \ref{fig:picture003}, namely, day high-density, day low-density, night high-density and night low-density. From which we know that: Under the four complex scenarios with different modes, the predicted trajectory of agents of our model is the most accurate, whether the agent is pedestrians, vehicles or riders.

\begin{figure}[!htbp]
\centering
\includegraphics[width=8.4cm]{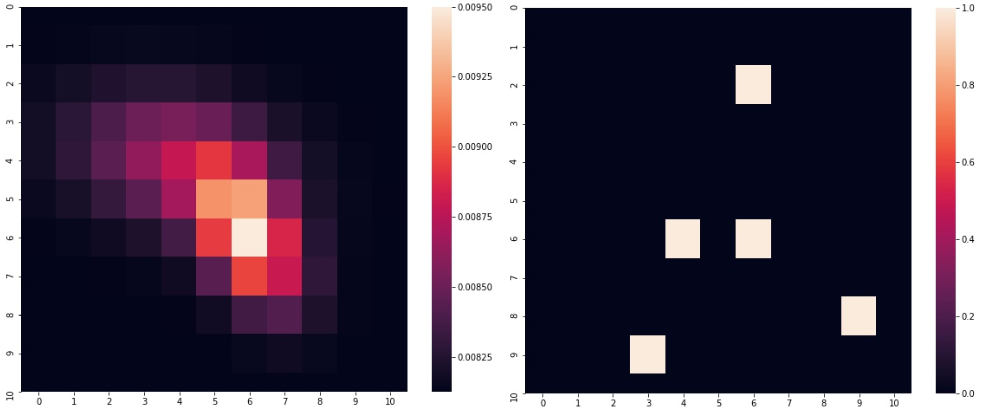}
\caption{The visualization of the attention mask.The left is the heat map of the attention mask, and the right is the distribution of agents in the grid cell. The cells with a value of 1 are the cells containing agents}
\label{fig:picture004}
\end{figure}
In addition, in order to visualize what our attention module learned, we visualized a typical mask as shown in Fig. \ref{fig:picture004}, it is worth noting that:
\begin{itemize}
  \item The proposed framework a simple and effective method to model the different types of traffic agents and use this method to improve the trajectory prediction precision.
  \item The high-weight regions in the attention mask learned by our model are strips. Considering that the driving routes of agents are also strips, we speculate that the model predicts which agents around are important according to its intention.
\end{itemize}
We draw the conclusion that our model uses agent centric's intention-based attention convolution fusion representation to improve the performance of trajectory prediction compared with existing methods.

\subsection{Ablation Study}
To discuss the impact of this information fusion mechanism on our model, we adopt the following two mechanisms to replace our convolution network (SCNN): concatenating directly (CON) and social pooling (SP).
The experimental results are shown in Tab. \ref{table3}, our model achieves the lowest prediction error, in which our encoder module variable-length LSTM and attention mask play an important role. Besides, the ConvNet plays a very important role in extracting the interactive information. Besides, from Tab. \ref{table4}, the longer horizon step will cause a decrease in prediction accuracy.

\begin{table}
\caption{The results of different combination models}
\begin{tabular*}{\hsize}{@{}@{\extracolsep{\fill}}cccc@{}}
\toprule
Dataset                     & \multicolumn{3}{c}{all} \\
\hline
Model \verb|\| Metric & ADE  & MDE  & FDE  \\
\midrule
VLSTM + CON                         &1.08	&1.77	&1.72\\
VLSTM + SP                         & 0.86	& 1.26	& 1.16 \\
VLSTM + SCNN                      & 0.82 	& 1.26  & 1.17  \\
LSTM+Attention+SCNN                     &  0.76	& 1.23 &	1.16\\
VLSTM+Attention+SCNN                        & \textbf{0.65} & \textbf{1.04} & \textbf{0.93} \\

\bottomrule
\label{table3}
\vspace{-0.5cm}
\end{tabular*}
\end{table}

\begin{table}
\caption{The results of different number of prediction horizon step}
\begin{tabular*}{\hsize}{@{}@{\extracolsep{\fill}}cccc@{}}
\toprule
Dataset                     & \multicolumn{3}{c}{all} \\
\hline
frame \verb|\| Metric & ADE  & MDE  & FDE  \\
\midrule
5 frame                        & \textbf{0.65} & \textbf{1.04} & \textbf{0.93} \\
7 frame & 0.78 & 1.11 & 1.10 \\
9 frame & 0.80 & 1.15 & 1.14 \\

\bottomrule
\label{table4}
\vspace{-0.5cm}
\end{tabular*}
\end{table}
\section{CONCLUSIONS}
In this paper, we propose a trajectory prediction model that involves an attention mechanism and a social map. By comparing
our method with several existing methods on the BLVD
dataset and analyzing the ablative experiment, we conclude that 1) It is significant for the trajectory prediction to consider the social relationship between the surrounding agents on the target agent. 2) The attention mechanism significantly contributes to accuracy improvement.

\section{Acknowledgements}
This work is supported by the National Science Foundation of China (No. 61790562, 61790563, 61773312)
\addtolength{\textheight}{-12cm}   









\bibliographystyle{ieeetr}
\bibliography{name}

\end{document}